\documentclass{article}

\usepackage{microtype}
\usepackage{graphicx}
\usepackage{subcaption}
\usepackage{booktabs} %

\usepackage{hyperref}

\usepackage[accepted]{icml2026}

\usepackage{amsmath}
\usepackage{amssymb}
\usepackage{mathtools}
\usepackage{amsthm}
\usepackage{multirow}
\usepackage{colortbl}
\usepackage{tabularx}
\usepackage{xcolor}
\usepackage{tabularx}
\usepackage{flushend}
\usepackage[capitalize,noabbrev]{cleveref}

\theoremstyle{plain}

\theoremstyle{definition}

\theoremstyle{remark}

\usepackage[disable,textsize=tiny]{todonotes}

\icmltitlerunning{Mode Seeking meets Mean Seeking for Fast Long Video Generation}

\def\para#1{\vspace{0.25em}\noindent\textbf{#1.}}

\newcommand{\pf}[1]{\cellcolor{red!30}#1}
\newcommand{\ps}[1]{\cellcolor{orange!30}#1}
\newcommand{\pt}[1]{\cellcolor{yellow!30}#1}

\begin{document}

\twocolumn[
  \icmltitle{Mode Seeking meets Mean Seeking for Fast Long Video Generation}

  \icmlsetsymbol{equal}{*}

  \begin{icmlauthorlist}
    \icmlauthor{Shengqu Cai}{stanford,nvidia}
    \icmlauthor{Weili Nie}{equal,nvidia}
    \icmlauthor{Chao Liu}{equal,nvidia}
    \icmlauthor{Julius Berner}{nvidia}
    \icmlauthor{Lvmin Zhang}{stanford}
    \icmlauthor{Nanye Ma}{nyu}
    \icmlauthor{Hansheng Chen}{stanford}
    \icmlauthor{Maneesh Agrawala}{stanford}
    \icmlauthor{Leonidas Guibas}{stanford}
    \icmlauthor{Gordon Wetzstein}{stanford}
    \icmlauthor{Arash Vahdat}{nvidia}
  \end{icmlauthorlist}

  \vspace{0.25em}
  \centerline{\url{https://primecai.github.io/mmm/}}
  \vspace{0.25em}

  \icmlaffiliation{stanford}{Stanford University, California, USA}
  \icmlaffiliation{nvidia}{NVIDIA Research, California, USA}
  \icmlaffiliation{nyu}{NYU Courant, New York, USA}

  \icmlcorrespondingauthor{Shengqu Cai}{shengqu@cs.stanford.edu}

  \icmlkeywords{Machine Learning, ICML}

  \vskip 0.3in
]

\printAffiliationsAndNotice{\icmlEqualContribution}  %

\begin{abstract}
\looseness=-1
Scaling video generation from seconds to minutes faces a critical bottleneck: while short-video data is abundant and high-fidelity, coherent long-form data is scarce and limited to narrow domains.
To address this, we propose a training paradigm where Mode Seeking meets Mean Seeking, decoupling local fidelity from long-term coherence based on a unified representation via a Decoupled Diffusion Transformer~\cite{wang2025ddt}.
Our approach utilizes a global Flow Matching head trained via supervised learning on long videos to capture narrative structure, while simultaneously employing a local Distribution Matching head that aligns sliding windows to a frozen short-video teacher via a mode-seeking reverse-KL divergence.
This strategy enables the synthesis of minute-scale videos that learns long-range coherence and motions from limited long videos via supervised flow matching, while inheriting local realism by aligning every sliding-window segment of the student to a frozen short-video teacher, resulting in a few-step fast long video generator.
Automatic evaluations, human ratings, and one-minute stress tests show that our method effectively closes the fidelity–horizon gap by jointly improving local sharpness/motion and long-range consistency.
\end{abstract}

\section{Introduction} \label{sec:intro}
\looseness=-1
The generative frontier is rapidly advancing from static images to the dynamic, complex realm of videos.
An important goal is to synthesize arbitrarily long, coherent, and high-fidelity video streams that mirror the rich temporal structure of our world.
Such long-horizon generation could enable applications like interactive world modeling for embodied agents and games, long-form story/film generation with persistent characters and scenes, and controllable video editing or animation that maintains identity and style over extended time.
Recent progress in diffusion models and transformers has yielded remarkable success in generating short video clips, typically lasting only a few seconds~\cite{wang2025wan, hong2023cogvideo, yang2025cogvideox, tencent2025hunyuan1_5}.

\begin{figure}[t]
\centering
\includegraphics[width=0.9999\linewidth]{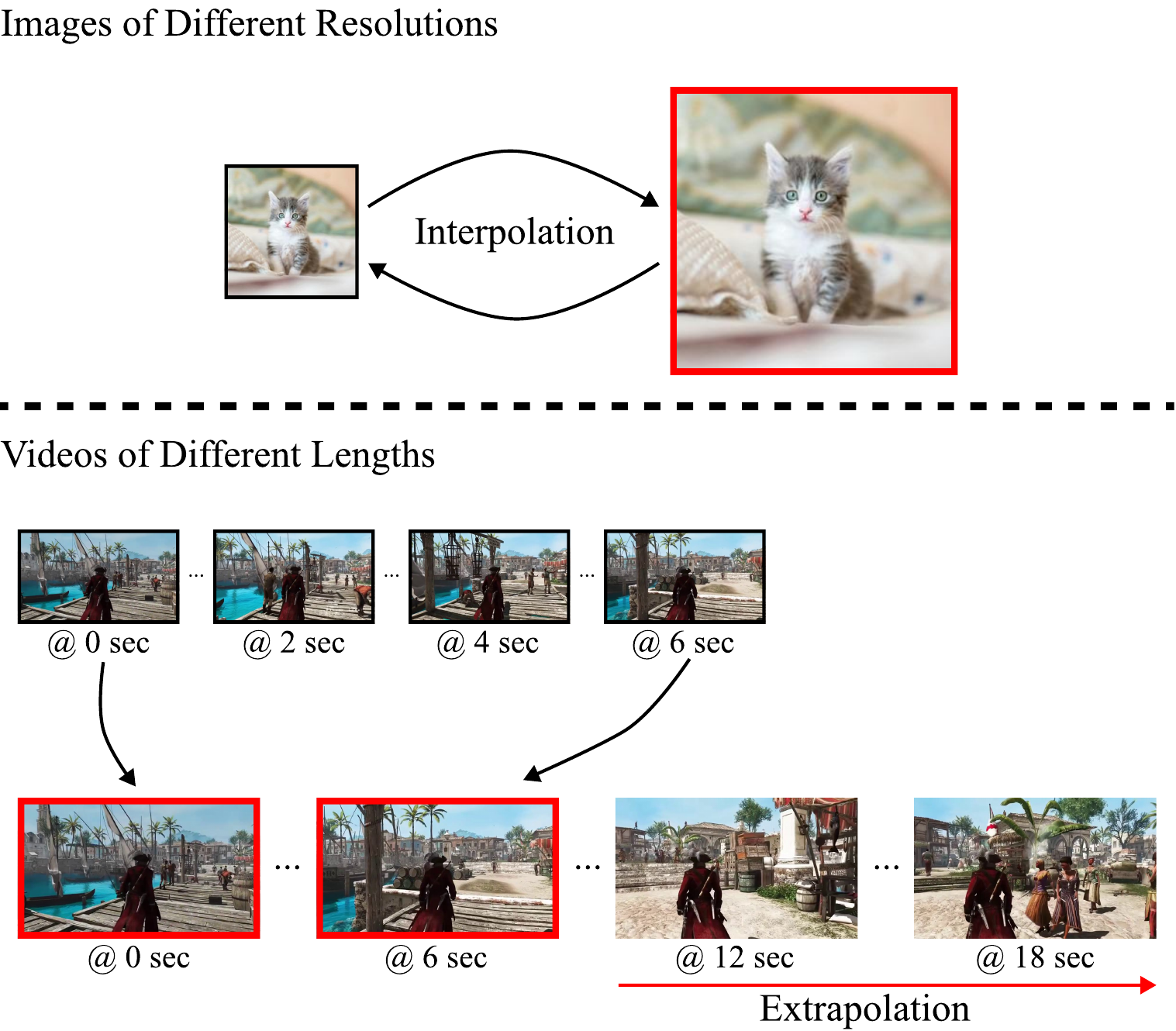}
\caption{
\looseness=-1
\textbf{Video length is not analogous to image resolution.}
\textit{Top: }For images, moving from low to high resolution is largely an interpolation of the same underlying local patch distribution. 
\textit{Bottom: }For videos, moving from short clips to long sequences is temporal extrapolation. The model must introduce new events and causal structure beyond the short-clip horizon, which is fundamentally harder than multi-resolution image training.
}
\label{fig:teaser}
\end{figure}

\looseness=-1
This success is enabled by a simple data reality: seconds-long clips are abundant online and thus available at web scale, e.g., in widely used video--text datasets~\cite{bain2021webvid, wang2024internvid, fan2025vchitect}.
Scaling from seconds to minutes breaks this recipe.
High-quality long-form videos that are minute-scale sequences with sustained events and context are far scarcer at a comparable scale, more heterogeneous, and substantially more expensive to curate and filter for generative training~(and post training)~\cite{li2024surveylongvideogeneration}.
As a result, the strongest available prior over realistic short-timescale dynamics often lives in a short-video generator---one that has already absorbed web-scale diversity and has been pushed toward high-fidelity modes through extensive refinement, while long-video training typically operates in a much more data-constrained regime and is extremely expensive.

\looseness=-1
A common practice borrowed from image generation is to train a single model on a ``soup'' of videos of varying lengths~\cite{yang2025cogvideox, kong2024hunyuan, tencent2025hunyuan1_5, chen2025skyreelsv2, seed2025seaweed7b, gao2025seedance1_0, seed2025seedance1_5, gao2025wans2v, bytedance2025contentv}.
The implicit hope is that the model will smoothly interpolate across temporal horizons, much as image models learn across spatial resolutions.
We argue that this assumption is fundamentally flawed, as shown in Fig.~\ref{fig:teaser}: the temporal dimension is not analogous to image resolution---a 1024$\times$1024 image is a higher-resolution interpolation of a 256$\times$256 image; its patches share essentially the same underlying distribution.
A one-minute video, however, is not an ``interpolation'' of a 5-second video: it is an ``extrapolation'' that adds new events, causal chains, and narrative structure, encoding substantially more information, in some sense, similar to panorama generation but requires significantly stronger context understanding ability.

\looseness=-1
This mismatch induces a critical and underappreciated failure mode.
Even when trained on mixed-duration data, a model that can generate longer sequences often comes at the expense of losing the sharp local dynamics characteristic in expert short-video teachers.
The outputs can look softer, less detailed, and less ``alive''.
In other words, the model is forced to relearn a high-fidelity short-video prior from a regime where the data and compute are most constrained, as evidenced in our experiment section.

We propose a training paradigm that decouples local fidelity from long-term coherence using a Decoupled Diffusion Transformer~\cite{wang2025ddt}~(DDT).
A long-video ``student" model learns global, minute-scale narrative structure through a mean-seeking supervised finetuning~(SFT) objective on limited long-video data.
Simultaneously, the student maintains local realism by aligning every generated sliding-window segment with a frozen, expert short-video teacher.
This alignment utilizes a mode-seeking reverse-KL divergence, allowing the student to inherit high-fidelity short-timescale priors by querying the teacher only on the student’s own rollouts.
To resolve the mathematical conflict between mean-seeking flow matching and mode-seeking distribution matching, we utilize two separate lightweight velocity heads that share a unified long-context encoder.
A Flow Matching head handles global coherence from real long videos, while a Distribution Matching head focuses on local realism distilled from the teacher.
At inference time, the Distribution Matching head serves as a fast, few-step sampler that produces videos with both sharp local dynamics and consistent long-range context.

\looseness=-1
Putting these pieces together, we close the fidelity--horizon gap with a simple decoupling: scarce long videos supervise global minute-scale structure, while a frozen expert short-video teacher provides a prior for local realism on every sliding window.
Moreover, because our teacher alignment is implemented through DMD~\cite{yin2024dmd, yin2024dmd2}-style distillation with a mode-seeking reverse-KL, the resulting DM head serves as a few-step sampler, enabling fast long-video synthesis at inference time.
\looseness=-1
We summarize our key contributions as follows:
\begin{itemize}
    \item We align every sliding-window segment of a long-video student to a frozen short-video teacher via a mode-seeking reverse KL, without requiring any additional short-video data.
    \item We use a DDT~\cite{wang2025ddt} with separate Flow Matching (long-video supervised finetuning) and Distribution Matching (teacher matching) heads, and show that such decoupled targets, decoded from a unified intermediate representation, are helpful for aligning long-context and local-quality targets.
    \item By using only the DM head for inference, we unlock fast, few-step long-video inference. Experiments show improved local quality/motion while preserving long-range consistency.
\end{itemize}
\looseness=-1

\section{Related Work}
\label{sec:related_work}
\para{Long video generation}
\looseness=-1
Diffusion dominates modern video synthesis, but scaling to long horizons remains challenging~\cite{zhang2026worldinworld, guo2024animatediff, guo2025lct}.
Training-free length extrapolation stretches pretrained models beyond their training horizon via noise rescheduling or temporal-frequency rebalancing~\cite{qiu2024freenoise, lu2024freelong, lu2025freelongpp, zhao2025riflex, ruhe2024rollingdiffusion}.
A complementary line blends diffusion with causal prediction---via noise-injected autoregressive rollouts, long-context AR designs~(e.g., flexible RoPE~\cite{su2021rope}), or teacher-to-student distillation---and appears in both research systems and production-scale deployments~\cite{chen2024diffusionforcing, song2025historyguidedvideodiffusion, gu2025far, su2021rope, yin2025causvid, kodaira2026streamdit, henschel2025streamingt2v, chen2025skyreelsv2, sandai2025magi1, guo2025selfresampling, po2025bagger}.
To mitigate AR drift, rollout-aware training~\cite{huang2025selfforcing} and extensions further improve length generalization and real-time rollout quality~\cite{huang2025selfforcing, cui2026selfforcingpp, liu2026rollingforcing, yang2026longlive, yi2026deepforcing, yesiltepe2026infinityrope, lu2025rewardforcing, chen2026contextforcing, lv2026lightforcing, wu2026infiniteworld}.

\para{Context learning and compression}
\looseness=-1
Long-context persistence is crucial as generation extends beyond a few seconds.
Retrieval-based memories ground prediction in relevant history (Field-of-View-, geometry-, or view-indexed)~\cite{xiao2025worldmem, yu2025contextasmemory, li2025vmem, huang2025memoryforcing, wu2025packandforce, fu2026plenoptic, xiao2025video4spatial, huang2026cinescene}, while learned routers/policies sparsify attention by selecting salient context chunks or token groups~\cite{cai2026moc, jia2026moga, meng2026holocine, wang2026multishotmaster, zhang2026blockvid, zhan2025bsa, zhang2025storymem, wen2026mitaattention, xie2025xstreamer, zhang2026sla2, zhang2026spargeattention2}.
Orthogonally, history can be compressed into a fixed-size state via latent packing or recurrent/state-space dynamics, with lightweight test-time adaptation providing another learned context representation~\cite{zhang2025framepack, xiao2026captaincinema, savov2025statespacediffuser, po2025ssmworldmodel, dalal2025ttt, zhang2026lact, zhang2025tinyhistory, ji2025memflow, li2026stablevideoinfinity, yu2025videossm}.

\para{Efficient video diffusion designs}
\looseness=-1
Large spatiotemporal contexts are compute-bound, motivating kernel-level optimizations such as FlashAttention~\cite{dao2022flashattention, dao2024flashattention2} and structured sparsity (e.g., sliding/tiling windows or radial masks) with training- or inference-time pruning~\cite{zhang2025sta, li2025radialattention, xi2025svg, yang2025svg2, xia2025adaspa, zhang2025jenga}.
Learned sparse routing/selection further keeps only salient token pairs or blocks~\cite{wu2026vmoba, zhang2025vsa, zhang2025sageattention, zhang2025sageattention2, zhang2025spargeattn}, while complementary directions reduce token/latent size or adopt multiscale and linear/block-linear attention to control memory growth~\cite{bolya2023tome, lee2024tokenmergevid, bachmann2025flextok, hacohen2024ltx, jin2025pyramidflow, xie2025sana, chen2026sanavideo}.

\para{Diffusion distillation}
\looseness=-1
Few-step diffusion distillation splits along the divergence used to align student and teacher: mean-seeking trajectory regression matches deterministic teacher targets via halved-step composition, PF-ODE self-consistency, average velocity, step-size conditioning, or rectified-flow straightening~\cite{salimans2022pd, song2023cm, luo2023lcm, lu2025scm, geng2025meanflow, frans2025shortcut, liu2023rectifiedflow, liu2024instaflow, tong2025freeflow}, while mode-seeking distribution matching minimizes a reverse-KL objective with score-difference gradients~\cite{poole2023dreamfusion, wang2023vsd, luo2023diffinstruct, zhou2024sid, sauer2024add} and underpins DMD-style image~\cite{yin2024dmd, yin2024dmd2} and video~\cite{yin2025causvid, huang2025selfforcing} distillation.
On highly multimodal distributions such as long videos, isotropic regression in pixel or weakly regularized latent space collapses modes and blurs high-frequency detail, motivating hybrids that interpolate the $f$-divergence family~\cite{xu2025fdistill}, replace MSE with a learned discriminator~\cite{lin2026cafm}, or fuse score distillation with consistency models~\cite{zheng2026rcm}.
We adopt a different decomposition: instead of blending the two divergences on a single output, we route them to disjoint content axes, using mean-seeking flow matching for global long-video structure and mode-seeking alignment for local sliding windows.

\begin{figure*}[t]
\centering
\includegraphics[width=0.9999\linewidth]{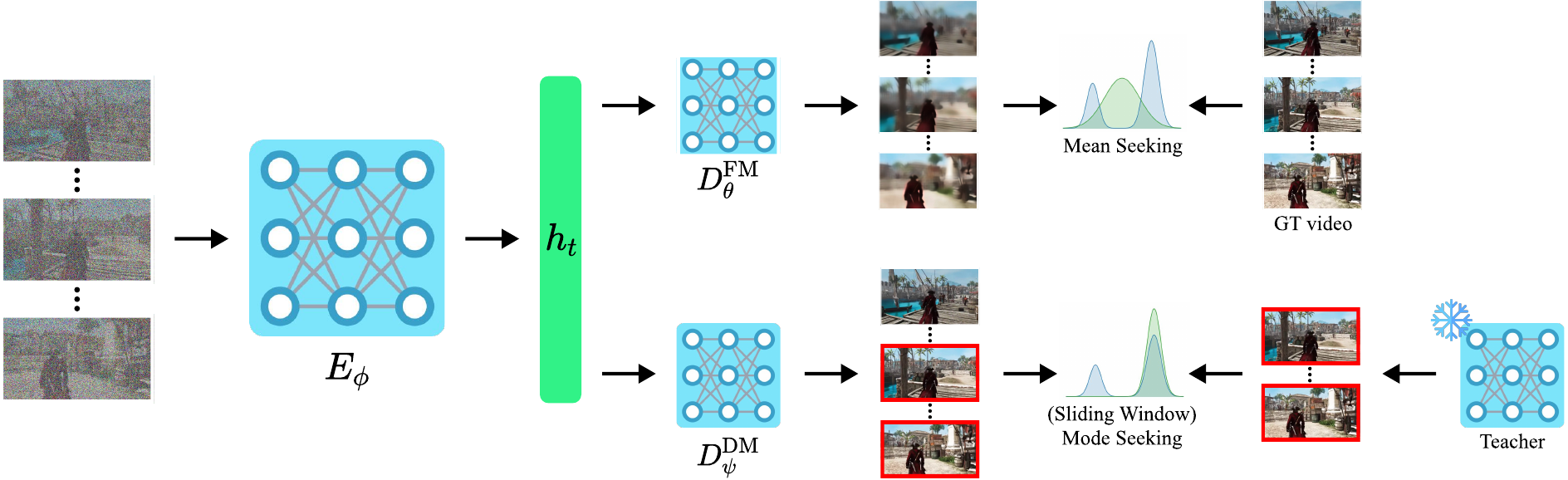}
\caption{\textbf{Overview: mode seeking meets mean seeking.}
A shared long-context condition encoder $E_\phi$ maps a noisy long-video latent $x_t^{\text{long}}$ (with timestep $t$ and conditioning $c$) to a unified representation $h_t$.
Two lightweight decoder heads read out velocities from $h_t$: the long context Flow Matching head $D^{\text{FM}}_\theta$ is trained with supervised flow matching on real long videos~(mean-seeking), while the segment-wise Distribution Matching head $D^{\text{DM}}_\psi$ is trained via on-policy sliding-window reverse-KL alignment to an expert short-video teacher using DMD~\cite{yin2024dmd, yin2024dmd2}/VSD~\cite{wang2023vsd}-style gradients~(mode-seeking).
Both objectives update the shared encoder, but each head receives only its corresponding signal.
}
\label{fig:overview}
\end{figure*}

\section{Method}
\label{sec:method}
\subsection{Preliminaries: Rectified Flow for Video Generation}
\label{sec:prelim}
\looseness=-1
We work in the latent space of a video VAE and follow the rectified flow parameterization~\cite{lipman2023flowmatching, liu2023rectifiedflow} used in recent video models.
Let $x_0 \in \mathbb{R}^{T \times H \times W \times C}$ denote a clean video latent (short or long), and let $z \sim \pi \triangleq \mathcal{N}(0, I)$ be a prior sample.
We construct a deterministic noising path
\begin{equation}
x_t \;=\; I_t(x_0, z) \;\triangleq\; (1 - t)\,x_0 + t\,z, \quad t \in [0,1],
\label{eq:linear-interp}
\end{equation}
which induces a corresponding generative ordinary differential equation~(ODE)
\begin{equation}
\frac{d x_t}{dt} = -u(x_t, t),
\qquad x_1 \sim \pi,
\label{eq:ode}
\end{equation}
where $u : \mathbb{R}^d \times [0,1] \to \mathbb{R}^d$ is the marginal velocity field and $d$ is the latent dimensionality.
\looseness=-1
A standard flow-matching objective trains a network $u_\theta$ to approximate $u$ via
\begin{equation}
\begin{aligned}
\mathcal{L}_{\text{FM}}(\theta)
\;=\;
\mathbb{E}_{x_0, z, t}
\,\big\|
u_\theta(x_t, t, c) - u(x_t, t \mid x_0, z)
\big\|_2^2,
\end{aligned}
\label{eq:fm}
\end{equation}
where $u(x_t, t \mid x_0, z) = x_0 - z$, and $c$ denotes conditioning (e.g., text prompt).
Sampling then corresponds to solving~\eqref{eq:ode} backward from $t=1$ to $t=0$ using $u_\theta$.
\looseness=-1
In our setting, we assume the existence of an expert short-video teacher, e.g., a pre-trained 5-second model $u_{\text{teacher}}$ (e.g., a DiT~\cite{peebles2023dit}-style flow model whose velocity field $u_{\text{teacher}}(x_t,t,c)$ is available at arbitrary $(x_t,t,c)$, but whose original short-clip training data and post-training strategies need not be accessible.

\subsection{Sliding-window View of Long Videos}
\looseness=-1
Given a long video latent $x_0^{\text{long}}$ of $T_{\text{long}}$ frames and a window length $L$ (corresponding to $\approx 5$\,s), we denote by
\begin{equation}
\mathrm{crop}_k(x_0^{\text{long}}) \in \mathbb{R}^{L \times H \times W \times C},
\end{equation}
the contiguous $L$-frame segment starting at frame index $k \in \{0, 1,\dots, T_{\text{long}} - L\}$.
Let $q_\Phi^{(k)}$ denote the marginal distribution over such windows induced by the long-video student model with parameters $\Phi$, and let $p_{\text{teacher}}$ denote the short-video teacher distribution.
Our goal is to regularize
\begin{equation}
\mathcal{L}_{\text{seg}}(\Phi)
\;=\;
\mathbb{E}_{k}\big[
D_{\mathrm{KL}}\!\big(q_\Phi^{(k)} \,\|\, p_{\text{teacher}}\big)
\big],
\label{eq:seg-kl}
\end{equation}
i.e., every sliding window produced by the long-video generator should match the short-video teacher distribution.
Because we use the reverse KL, $D_{\mathrm{KL}}(q \,\|\, p)$, this objective is mode-seeking: it encourages the student to concentrate its mass on the teacher's high-fidelity modes rather than averaging over them~\cite{yin2025causvid, huang2025selfforcing}. It in addition views a long generator as inducing sliding-window marginals and enforces Eq.~\eqref{eq:seg-kl} on windows sampled on-policy from long rollouts, using the short teacher as a local realism critic beyond its native horizon.
\subsection{Decoupled Long-Video Diffusion Transformer}
\label{sec:ddt-arch}
\looseness=-1
When viewing the reverse KL term in Eq.~\eqref{eq:seg-kl} as an alignment target, it introduces a fundamental tension with the standard flow-matching objective in Eq.~\eqref{eq:fm}.
Denote by $f(x_t) = x_t + t\,u(x_t,t,c)$ the per-step clean-video prediction induced by a velocity head $u$ (Eq.~\eqref{eq:linear-interp}).
At a fixed noise level $t>0$, the pointwise optimum of the squared flow-matching/SFT objective is mean-seeking:
$f_{\mathrm{SFT}}^*(x_t)=\mathbb{E}_{p_{\text{long}}}[x_0\mid x_t, c]$.
For a multimodal posterior $p_{\text{long}}(x_0\mid x_t, c)$, which is typical at high noise levels, this expectation averages across diverse modes and can lie off the data manifold, producing blurry or unrealistic predictions.
Conversely, DMD aligns the distribution of predictions with the teacher through a mode-seeking reverse-KL objective, encouraging sharp, realistic outputs.
Applying both signals to a single velocity predictor creates competing pressures: long-video SFT favors conditional-average predictions, while teacher distribution matching favors sample-like predictions.
This motivates us to instantiate the long-video student as a decoupled architecture, where a feature encoder is followed by two decoder heads, inspired by DDT~\cite{wang2025ddt}.
An overview of our design is demonstrated in Fig.~\ref{fig:overview}.

\para{Condition encoder}
\looseness=-1
Given a noisy long video $x_t^{\text{long}}$, conditioning $c$, and timestep $t$, the encoder
\begin{equation}
h_t \;=\; E_\phi(x_t^{\text{long}}, t, c)
\end{equation}
produces a spatiotemporal feature tensor $h_t$.
Architecturally, $E_\phi$ is a video diffusion transformer with full-range temporal dependencies with full attention, and forms the shared backbone between all heads.

\para{Two velocity heads}
\looseness=-1
On top of $h_t$ we attach two lightweight transformer decoders:
\begin{align}
u_\theta(x_t^{\text{long}}, t, c)
&= D^{\text{FM}}_\theta(h_t, t, c),
\label{eq:fm-head}
\\
v_\psi(x_t^{\text{long}}, t, c)
&= D^{\text{DM}}_\psi(h_t, t, c).
\label{eq:dm-head}
\end{align}
The Flow Matching~(FM) head $D^{\text{FM}}_\theta$ parameterizes the student's global velocity field $u_\theta$ used for long-horizon training and sampling on ground truth long videos (Sec.~\ref{sec:sft-long}) using Eq.~\eqref{eq:fm}.
The Distribution Matching~(DM) head $D^{\text{DM}}_\psi$ is a few-step generator that uses teacher alignment on local windows.
\looseness=-1
Sharing the encoder $E_\phi$ but decoupling the heads has two advantages:
(i) the long-context representation $h_t$ is learned and reused across objectives, and
(ii) the short-video generation capability is distilled from the teacher distribution without forgetting when training on scarce long video data.
Next, we discuss how the two heads are trained.
\subsection{Local Reverse-KL via DMD/VSD}
\label{sec:dmd}
\looseness=-1
We now introduce our local distribution matching loss, which regularizes the student's local dynamics and quality using the short-video teacher.
Conceptually, we would like each window marginal $q_\Phi^{(k)}$~(where $\Phi \triangleq (\phi, \psi)$) of the student to match the teacher distribution $p_{\text{teacher}}$ via a mode-seeking reverse KL, as shown in Eq.~\eqref{eq:seg-kl}.
However, directly evaluating Eq.~\eqref{eq:seg-kl} is intractable.
We instead follow the DMD/VSD literature~\cite{yin2024dmd, yin2024dmd2, wang2023vsd}, which shows that for diffusion/flow models, the gradient of a reverse-KL between student and teacher can be expressed in terms of their score/velocity differences on noisy states.
We adapt this to videos with sliding local windows.

\para{Reverse-KL gradient on noised windows}
\looseness=-1
Let $q_{\Phi,t}^{(k)}$ denote the distribution of the noised window
$\hat{x}_t^{(k)} = (1-t)\,\hat{x}_0^{(k)} + t\,\epsilon$
where $\hat{x}_0^{(k)}$ is a sample from the student $q_{\Phi}^{(k)}$ and $\epsilon \sim \mathcal{N}(0, I)$ is fresh noise.
DMD shows that the reverse-KL gradient can be written as an expectation over the student distribution,
involving the difference between the teacher's score and the student's own (``fake'') score on the same noisy state.
Concretely, using the linear interpolation in Eq.~\eqref{eq:linear-interp} (where $\alpha_t = 1-t$),
we use the window-level gradient surrogate
\begin{equation}
\begin{split}
\widehat{\nabla}\, \mathcal{L}_{\text{seg}}
&=
\mathbb{E}_{\,t,\,k}
\Big[
\lambda(t)\,
\big(
v_{\text{fake}}(\hat{x}_t^{(k)}, t, c)
\\
&
-
u_{\text{teacher}}(\hat{x}_t^{(k)}, t, c)
\big)^{\!\top}
\nabla\, \hat{x}_0^{(k)}
\Big].
\end{split}
\label{eq:dmd-grad}
\end{equation}
where $\hat{x}_0^{(k)}$ is the generated student window, and $v_{\text{fake}}$ is a fake score estimator trained on the student's window predictions $\hat{x}_0^{(k)}$ using score matching for 5 steps in between the student updates.
We treat $(v_{\text{fake}}-u_{\text{teacher}})$ as a stop-gradient term and backpropagate only through $\hat{x}_0^{(k)}$.
Here $\lambda(t)$ absorbs the standard DMD/VSD weighting (including score-to-velocity scaling),
$u_{\text{teacher}}(\cdot)$ is the short-video teacher velocity query, and $v^{(k)}_{\psi}(\cdot)$ is the student's window velocity predicted by the mode-seeking head, obtained as follows.
\begin{table*}[t]
\centering
\resizebox{0.9999\linewidth}{!}{ %
\begin{tabular}{@{} l | c | c | c | c | c | c | c | c }
\toprule
\multirow{2}{*}{Method} & \multirow{2}{*}{NFE} & \hspace{12pt} Subject \hspace{5pt} \multirow{2}{*}{$\uparrow$} & \hspace{9pt} Background \hspace{2pt} \multirow{2}{*}{$\uparrow$} & \hspace{14pt} Motion \hspace{9pt} \multirow{2}{*}{$\uparrow$} & \hspace{3pt} Dynamic\multirow{2}{*}{$\uparrow$} & Aesthetic\multirow{2}{*}{$\uparrow$} & \hspace{9pt} Image \hspace{3pt} \multirow{2}{*}{$\uparrow$} & \hspace{15pt} VLM \hspace{9pt} \multirow{2}{*}{$\uparrow$} \\
& & Consistency & Consistency & Smoothness & Degree & Quality & Quality & Consistency \\
\midrule
Long-context SFT & 50 & \pt{0.9685} & 0.9533 & \ps{0.9866} & \ps{0.9375} & 0.4973 & 0.6303 & \pf{77.28} \\
Mixed-length SFT & 50 & 0.9667 & 0.9541 & \pf{0.9874} & 0.8906 & 0.5467 & 0.6683 & \pt{74.63} \\
\midrule
CausVid~\cite{yin2025causvid} & 4 & \pf{0.9736} & \pf{0.9614} & 0.9789 & 0.8594 & \pf{0.6044} & 0.6305 & 39.30\\
Self Forcing~\cite{huang2025selfforcing} & 4 & 0.9489 & 0.9451 & 0.9805 & \pt{0.9063} & \pt{0.5556} & 0.6278 & 37.60\\
InfinityRoPE~\cite{yesiltepe2026infinityrope} & 4 & \ps{0.9689} & \ps{0.9573} & 0.9812 & 0.7188 & 0.5342 & \ps{0.6871} & 68.61 \\
LongLive~\cite{yang2026longlive} & -- & 0.9681 & \pt{0.9565} & 0.9838 & 0.6422 & 0.5401 & \pt{0.6684} & 62.92 \\
\midrule
Ours & 4 & \underline{0.9682} & \underline{0.9548} & \pt{\underline{0.9863}} & \pf{\underline{0.9453}} & \ps{\underline{0.5735}} & \pf{\underline{0.6982}} & \ps{\underline{75.42}}\\
\bottomrule
\end{tabular}
} %
\caption{
\textbf{Quantitative comparisons.}
The \colorbox{red!30}{first}, \colorbox{orange!30}{second}, and \colorbox{yellow!30}{third} values are highlighted.
NFE is reported for diffusion/flow samplers in our setup; LongLive~\cite{yang2026longlive} is a frame-level AR real-time system, so its NFE is not directly comparable.
We note that rollout-based methods can get over-saturated~(CausVid~\cite{yin2025causvid}) or static/conservative~(InfinityRoPE~\cite{yesiltepe2026infinityrope}, LongLive~\cite{yang2026longlive}), which in turn significantly boosts some consistency metrics.
Even so, our method still stands out as the best overall performing model.
}
\label{tab:quantitative}  
\end{table*}

\para{Cropping the mode-seeking head}
\looseness=-1
We obtain this generated window by cropping the mode-seeking head's output on the noised long video latent sequences:
\begin{align}
x_t^{\text{long}} &= (1-t)\,x_0^{\text{long}} + t\,\epsilon,
\\
h_t &= E_\phi(x_t^{\text{long}}, t, c),
\\
v_{\psi}^{\text{long}}(x_t^{\text{long}}, t, c) &= D^{\text{DM}}_\psi(h_t, t, c),
\\
v^{(k)}_{\psi}(x_t^{\text{long}}, t, c) &= \mathrm{crop}_k\!\Big(v_{\psi}^{\text{long}}(x_t^{\text{long}}, t, c)\Big),
\end{align}
where $\hat{x}_t^{(k)} = \mathrm{crop}_k(\hat{x}_t^{\text{long}})$ and $\epsilon \sim \mathcal{N}(0,I)$.
The teacher term $u_{\text{teacher}}(\hat{x}_t^{(k)},t,c)$ is evaluated on the cropped $L$-frame window using the short-video teacher.
Sliding Window DMD~\cite{yin2024dmd} is non-trivial to implement on modern video generation models.
In particular, their latent space often includes both image latents and video frame latents to better support image-video joint training~\cite{wang2025wan, kong2024hunyuan}.
When applying DMD with sliding windows on long latent sequences, a window cropped from the middle begins with a video latent, while the teacher expects the first latent of a clip to be an image latent, leading to a semantic mismatch at window boundaries.
To address this, we follow LongLive~\cite{yang2026longlive}: for any window starting at offset $p>0$, we decode the latent prefix $[0,...,p-1]$ with the frozen VAE, take the last decoded RGB frame, and re-encode it into an image latent; we then prepend this reconstructed image latent to the student's windowed video latents before computing the DMD loss, masking out the reconstructed latent to avoid backpropagating through the VAE.
We find this strategy to be effective not only for causal AR models~\cite{yang2026longlive}, but also for non-causal bidirectional models like ours.
\begin{figure*}[t]
\centering
\includegraphics[width=0.9999\linewidth]{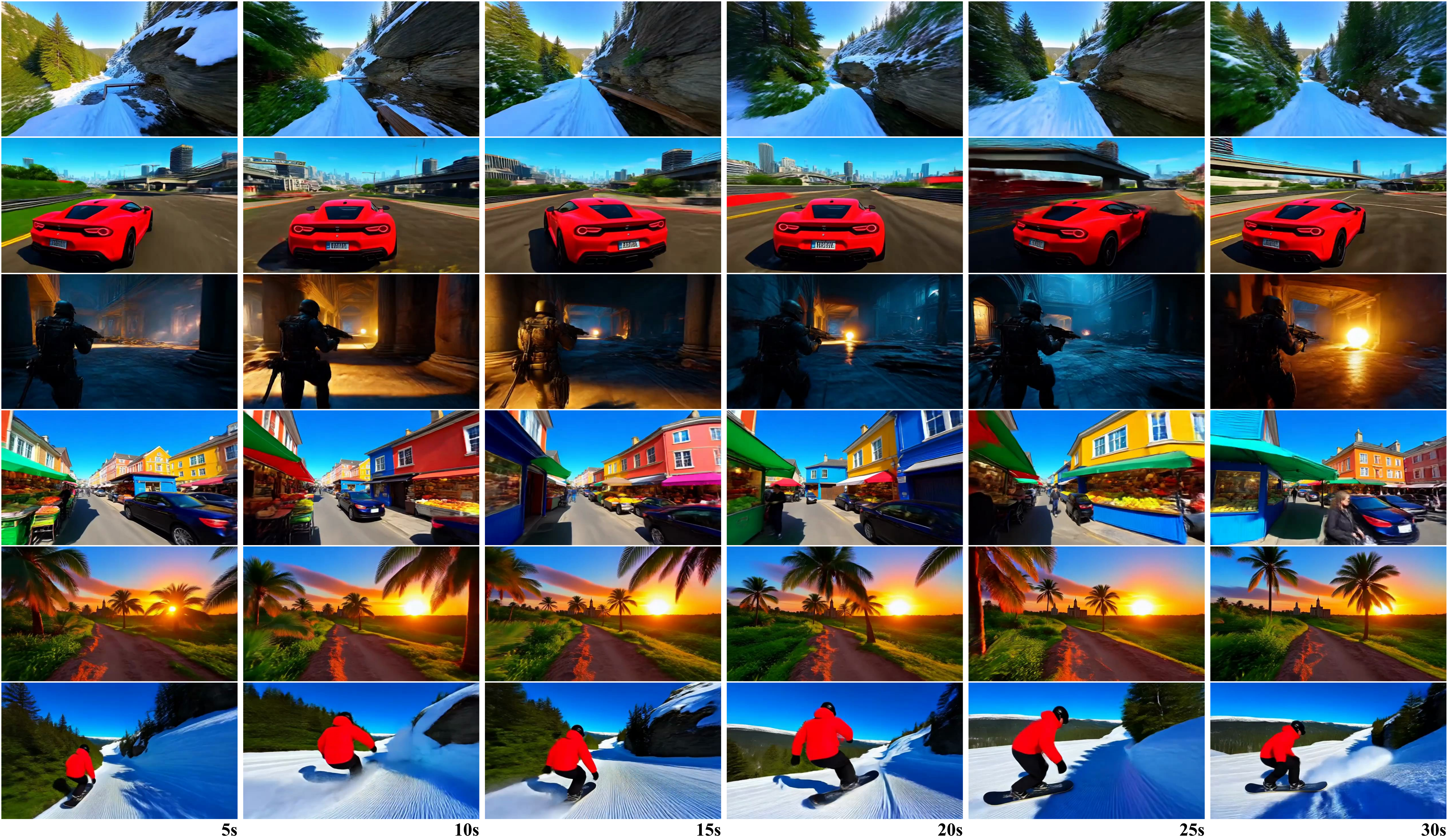}
\caption{\textbf{Qualitative results.}
Our method generalizes well to various scenarios, producing long videos that maintain local fidelity and global coherence.
All results are obtained using the Wan~\cite{wang2025wan} 1.3B model as both the student and the teacher, demonstrating how our decoupled training effectively extends short-video capabilities to long-horizon generation.
We refer to our supplemental website for the videos and results from the Wan~\cite{wang2025wan} 14B model.
}
\label{fig:qualitative_results}
\end{figure*}

\subsection{SFT Flow-Matching Anchor on Long Videos}
\label{sec:sft-long}
\looseness=-1
Distribution matching teacher alignment alone cannot teach global, minute-scale coherence: the short-video teacher does not model long-range structure by design.
To learn long-range dynamics from the limited long-video corpus, we train the FM head $u_\theta$ with a supervised flow-matching~(SFT) objective on full-length videos.
Let $x_0^{\text{long}} \sim p_{\text{long}}$ be a real long video latent, and let $x_t^{\text{long}}$ be constructed as in~\eqref{eq:linear-interp} using a global prior $z^{\text{long}} \sim \pi$.
We train $u_\theta$ with
\begin{equation}
\begin{split}
\mathcal{L}_{\text{SFT}}(\phi,\theta)
&=
\mathbb{E}_{x_0^{\text{long}}, z^{\text{long}}, t}
\Big[
\big\|
u_\theta(x_t^{\text{long}}, t, c)
\\
&\qquad\qquad
- \big(x_0^{\text{long}} - z^{\text{long}}\big)
\big\|_2^2
\Big],
\end{split}
\label{eq:sft-long}
\end{equation}
where gradients flow through both the condition encoder $E_\phi$ and the FM head $D^{\text{FM}}_\theta$.
This objective anchors the student's global velocity field to real long video trajectories, encouraging correct long-horizon temporal dependencies and narrative structure.
\subsection{Joint Objective and Training Procedure}
\label{sec:joint-objective}
\looseness=-1
Our training combines two complementary signals: (i) supervised long-video flow matching on the global FM head, and (ii) local reverse-KL alignment on the DM head via the DMD/VSD gradient surrogate.
Concretely, we optimize the long-video model with
\begin{equation}
\mathcal{L}_{\text{total}}(\phi,\theta,\psi)
\;=\;
\mathcal{L}_{\text{SFT}}(\phi,\theta)
\;+\;
\lambda_{\text{seg}}\;\mathcal{L}_{\text{seg}}(\phi,\psi),
\label{eq:total-loss}
\end{equation}
where $\lambda_{\text{seg}}$ is a scalar weight, $\mathcal{L}_{\text{SFT}}$ is the supervised FM objective on real long videos in Eq.~\eqref{eq:sft-long}, and $\mathcal{L}_{\text{seg}}$ denotes the conceptual window-level reverse KL in Eq.~\eqref{eq:seg-kl}.
Since $\mathcal{L}_{\text{seg}}$ is intractable to evaluate, we implement its effect by injecting the DMD/VSD gradient surrogate in Eq.~\eqref{eq:dmd-grad}.
The shared encoder $E_\phi$ receives gradients from both terms, while the two heads are updated by their respective signals:
\begin{align}
\nabla_{\phi}\mathcal{L}_{\text{total}}
&=
\nabla_{\phi}\mathcal{L}_{\text{SFT}}
+
\lambda_{\text{seg}}\;\widehat{\nabla}_{\phi}\mathcal{L}_{\text{seg}},
\\
\nabla_{\theta}\mathcal{L}_{\text{total}}
&=
\nabla_{\theta}\mathcal{L}_{\text{SFT}},
\\
\nabla_{\psi}\mathcal{L}_{\text{total}}
&=
\lambda_{\text{seg}}\;\widehat{\nabla}_{\psi}\mathcal{L}_{\text{seg}}.
\end{align}
This realizes the intended decoupling: long-video supervision updates the global FM head, while local short-video teacher distribution matching is handled by the DM head.
At the same time, they both act through and update the shared representation.
Each step uses two minibatches:
(1) a batch of real long videos to compute $\mathcal{L}_{\text{SFT}}$ and update $(\phi,\theta)$;
(2) a batch of on-policy student rollouts to sample windows and apply the DMD/VSD surrogate update in Eq.~\eqref{eq:dmd-grad} to $(\phi,\psi)$.
\subsection{Inference}
\label{sec:inference}
\looseness=-1
At inference time, we discard the FM head and generate long videos using the DM head $v_\psi$.
We note that the DM head, while supervised only on sliding windows similar to APT2~\cite{lin2025apt2}, can directly generate long videos end-to-end bidirectionally.
The shared encoder ensures that the representation used by $v_\psi$ has been shaped as follows: any sliding window of the generated long video must reside in the same local distributional modes as the expert short-video teacher, while long-range coherence and narrative structure are learned directly from scarce long videos via $\mathcal{L}_{\text{SFT}}$.

\looseness=-1
Aforementioned decoupled design allows the model to simultaneously optimize for local realism and long-horizon coherence:
the short-video teacher provides a mode-seeking regularizer on every local segment, and the long-video SFT objective concentrates the limited long-form data and capacity on what the teacher cannot provide: minute-scale temporal coherence.
In addition, we no longer need multi-stage training and distillation; the output model with DM head is directly capable of few-step inference, unlocking fast minute-scale video generation.

\section{Results}
\label{sec:results}
\subsection{Experimental Details}

\para{Base models}
\looseness=-1
We trained our methods on both the Wan~\cite{wang2025wan} 1.3B model and the Wan~\cite{wang2025wan} 14B model.
For fair comparison, we use Wan~\cite{wang2025wan} 2.1 1.3B throughout our quantitative and qualitative comparisons, both as the student and as the teacher whenever involved.
We refer to our supplementary website for qualitative results on the 14B model.

\para{Data}
\looseness=-1
We collect and use data from several sources.
From publicly available datasets, we use all videos available from the Sekai dataset~\cite{li2025sekai}, and a subset from MiraData~\cite{ju2024miradata}.
We also filter out single-shot videos from randomly collected internet videos.
These datasets include more than 100k videos ranging from 10 seconds to minutes, with an average of 31 seconds.
We set the temporal upper bound to 61 seconds and subsample when a video exceeds it.

\para{Implementation details}
\looseness=-1
We train our model on A100 and GB200 GPUs.
We use dynamic batching, treating training videos as variable-length sequences and pre-processing them into length-based buckets for sampling.
Variable-length training is used across training, reducing padding waste and IDLE time.
To train for long contexts, we use the DeepSpeed Ulysses~\cite{jacobs2023ulysses} sequence-parallelism strategy.
We use a sequence-parallelism group size of 4 for A100 GPUs, and 2 for GB200 GPUs.
We implemented everything on top of the FastGen~\cite{nie2026fastgen} repository.

\para{Baselines}
\looseness=-1
We set up the following two baselines that are widely used in modern long video models/video world models:
\textit{(1) Long-context SFT}: the very basic long tuning strategy, where long video clips are collected and used to finetune the short pretrained model;
\textit{(2) Mixed-lengths SFT}: the advanced setup used by industrial-level model training, when videos of different lengths are jointly trained, hoping to achieve interpolation among modes of different lengths.
We also compare our method against popular auto-regressive video generation methods, CausVid~\cite{yin2025causvid}, Self-Forcing~\cite{huang2025selfforcing}, where we use the extrapolation method provided in CausVid to generate long videos.
Additionally, we compare with a popular Self-Forcing extrapolation method, InfinityRoPE~\cite{yesiltepe2026infinityrope}.
We further include LongLive~\cite{yang2026longlive}, a recent real-time frame-level autoregressive long-video generator.
We note that our method is orthogonal to causal AR models and can be used alongside them, which is an interesting follow-up.

\begin{figure*}[t!] %
    \centering        %
    \includegraphics[width=0.9999\linewidth]{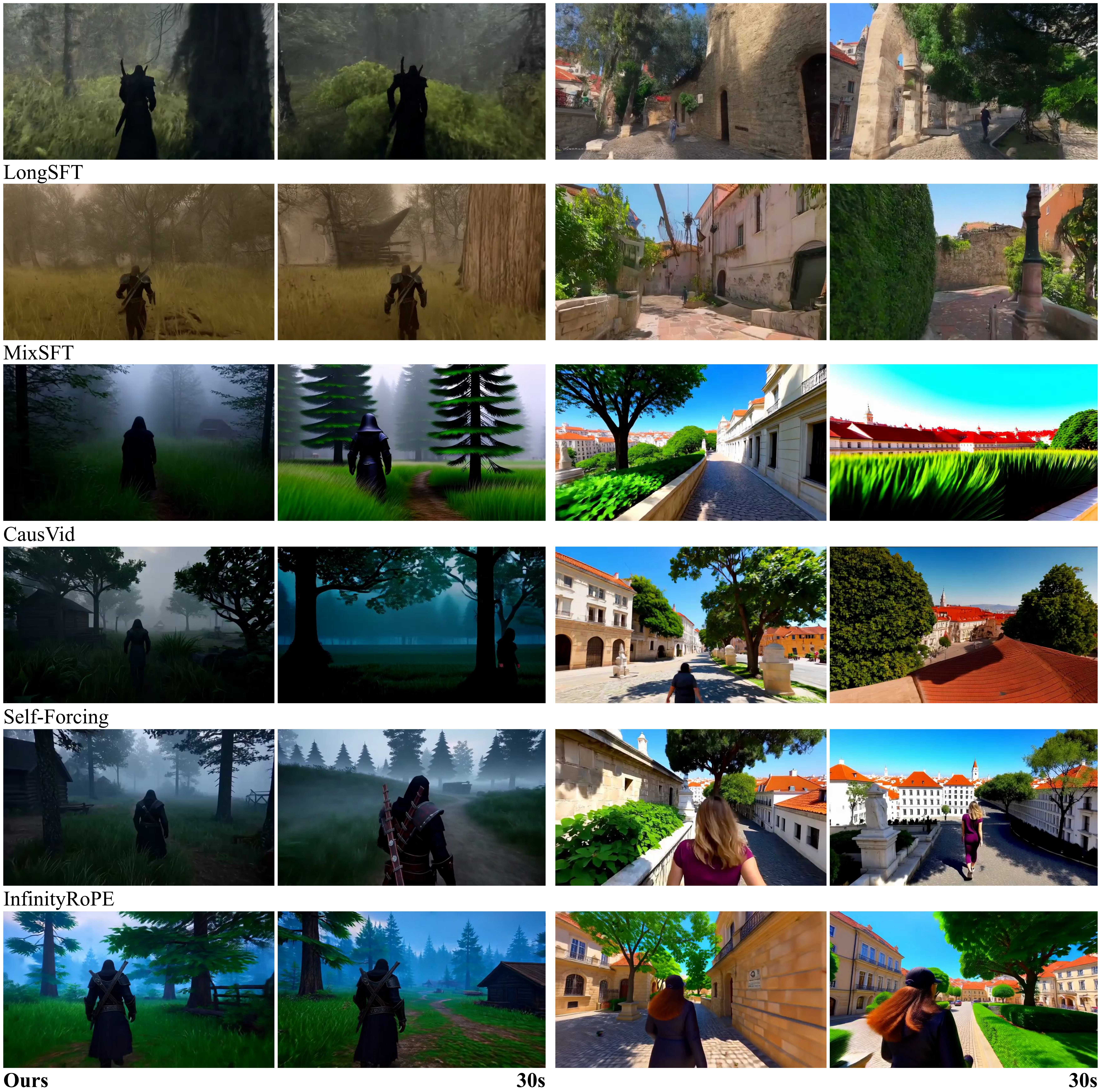}
    \caption{\textbf{Qualitative comparison.}
    ``LongSFT'' and ``MixSFT'' refer to long-context supervised finetuning~(SFT) and mixed-length SFT, respectively.
    SFT-only methods~(LongSFT, MixSFT) achieve decent long context and narrative, but appear to be blurry.
    Teacher-only methods~(CausVid~\cite{yin2025causvid}, Self-Forcing~\cite{huang2025selfforcing}) suffer from quality degradation over long range.
    While InfinityRoPE~\cite{yesiltepe2026infinityrope} extends the temporal horizon, it does not have the ability to process long context and tends to generate still contents.
    Our method stands out as the best-performing model overall in terms of quality, motion, and long-horizon consistency.
    We refer to our supplementary website for video results, including extended comparisons with LongLive~\cite{yang2026longlive}.
    }
    \label{fig:qualitative_comparison}
    \vspace{-5pt}  %
\end{figure*}

\para{Evaluation metric}
We use 200 prompts describing long videos and events as our test set, and generate 30-second videos.
We adopt the metrics and evaluation protocol used in VBench-Long~\cite{huang2024vbench, huang2026vbench++, zheng2025vbench2}, which includes subject consistency, background consistency, motion smoothness, temporal flickering, dynamic degree, aesthetic quality, and imaging quality.
We also report consistency evaluated by Gemini-3-Pro, an advanced MLLM capable of taking video inputs.
The Gemini instruction prompt is provided in our supplementary Sec.~\ref{sec:gemini_eval_prompt}.

\subsection{Quantitative Results}
\looseness=-1
We present the quantitative results in Tab.~\ref{tab:quantitative}, and provide analysis for each class of methods in the following sections.
We additionally include 60-second automatic results and human ratings in the supplementary material~(Sec.~\ref{sec:one_minute_generation} and Sec.~\ref{sec:user_study}).

\para{Analysis on SFT-only methods}
\looseness=-1
Supervised Fine-Tuning~(SFT), particularly the Mixed-Length SFT strategy used in industrial models, remains a robust baseline for establishing temporal coherence.
By training jointly on videos of varying lengths, the model learns to interpolate and extend motion patterns beyond the short-clip horizon.
However, the effectiveness of SFT is fundamentally capped by the scarcity of high-quality training data.
While short video clips are abundant, high-resolution, single-shot long videos~(e.g., $>30$ seconds) with sustained narrative continuity are extremely rare.
Thus, while SFT is important for learning long-context global structure, it is insufficient for maintaining high fidelity across diverse domains due to data gaps, as evidenced by their relatively low quality scores.

\para{Analysis on teacher-only methods}
\looseness=-1
Methods that rely exclusively on distilling a short-video teacher or rolling out a local generator~(e.g., CausVid~\cite{yin2025causvid}, Self-Forcing~\cite{huang2025selfforcing}, InfinityRoPE~\cite{yesiltepe2026infinityrope}, and LongLive~\cite{yang2026longlive}) face a different set of challenges when scaled to minute-level generation. While these approaches succeed in ensuring local realism for short durations~($\approx 5$ seconds), they lack a mechanism to ground long-term narrative structure.
First, because these methods typically employ autoregressive~(AR) rollouts, they suffer from error accumulation; slight deviations in early frames compound over dozens of autoregressive steps, leading to significant quality degradation in minute-scale outputs.
Even with extension techniques like InfinityRoPE~\cite{yesiltepe2026infinityrope}, the fundamental issue remains: the model is trained without any ground-truth long-video supervision.
Critically, a short-video teacher is strictly ``blind'' to long-context concepts.
We also find ``sink''- or cache-based rollout methods can become conservative over long horizons, shown by the dynamic degree scores and by the static-artifact ratings in Sec.~\ref{sec:user_study}.

\para{Analysis on our method}
\looseness=-1
These observations validate our decoupled design.
By utilizing SFT on the global Flow Matching head, we provide the necessary ``long-context anchor'' that allows the model to understand long-scale prompts and temporal dependencies—capabilities that the short teacher cannot provide.
Simultaneously, by applying Teacher Distribution Matching exclusively via the mode-seeking head on local sliding windows, we inject the high-fidelity texture and motion priors of the teacher without requiring the teacher to understand long-range context.
\begin{table}[t]
\centering
\resizebox{0.9999\linewidth}{!}{ %
\begin{tabular}{@{} l | c | c | c }
\toprule
Method & Consistency $\uparrow$ & Motion $\uparrow$ & Quality $\uparrow$
\\
\midrule
No DDT dual heads & 0.9427 & 0.9449 & 0.5298 \\
No Sliding-window DMD & \ps{0.9604} & \pt{0.9621} & \ps{0.6075} \\
No SFT & \pt{0.9579} & \pf{0.9690} & \pt{0.5862} \\
\midrule
Full Model & \pf{0.9615} & \ps{0.9685} & \pf{0.6359} \\
\bottomrule
\end{tabular}
} %
\caption{
\textbf{Ablation study.}
The \colorbox{red!40}{first}, \colorbox{orange!50}{second}, and \colorbox{yellow!50}{third} values are highlighted.
Our full model achieves the best overall performance, highlighting the necessity of all components.
}
\vspace{-5pt}
\label{tab:ablation}  
\end{table}

\subsection{Qualitative Results}
\looseness=-1
We showcase qualitative results in Fig.~\ref{fig:qualitative_results}, and present comparisons with the baselines using representative frames in Fig.~\ref{fig:qualitative_comparison}.
We refer to our supplemental material for dynamic video comparisons.
Long-context SFT~(LongSFT) and Mixed-length SFT~(MixSFT) typically preserve the rough scene layout over time, but the outputs exhibit a characteristic loss of local realism: fine textures are washed out, edges appear softened, and the foreground subject can collapse into an under-defined silhouette (top example).
In the street-walk example~(bottom), SFT baselines also show signs of weakened long-range camera/scene continuity, with noticeable viewpoint/scene inconsistencies across frames.
These failure modes are consistent with the intuition that, under scarce long-video supervision, a single mean-seeking objective tends to average away high-frequency detail even within short temporal windows.
Teacher-only or rollout-based methods, such as CausVid~\cite{yin2025causvid}, Self-Forcing~\cite{huang2025selfforcing}, InfinityRoPE~\cite{yesiltepe2026infinityrope}, and LongLive~\cite{yang2026longlive}, can inherit stronger local contrast from the short-video teacher, but often struggle to maintain realistic evolution over extended horizons.
This appears to drift toward over-saturation and reduced dynamics, or conservative motion to avoid autoregressive error accumulation, which can manifest as visually stable yet less eventful sequences.
The motion degradation is especially notable in InfinityRoPE~\cite{yesiltepe2026infinityrope}, which uses "sink'' mechanisms to attach generation to certain frames.
Our method, by decoupling global mean-seeking supervision~(SFT flow matching on long videos) from local mode-seeking alignment~(reverse-KL distribution matching on sliding windows), simultaneously maintains long-range scene/narrative consistency and preserves crisp local appearance.
Concretely, the foreground subject remains sharper and more consistently rendered across time, while backgrounds evolve smoothly without the abrupt viewpoint shifts seen in SFT-only training or the motion collapse/pattern repetition commonly induced by teacher-only long rollouts.
Overall, Fig.~\ref{fig:qualitative_comparison} qualitatively supports our design goal: mode seeking meets mean seeking yields long-video generation that retains short-horizon realism while extending temporal coherence.

\subsection{Ablation}
\looseness=-1
As shown in Tab.~\ref{tab:ablation}, we ablate three key ingredients of our approach: (i) the decoupled DDT dual-head design, (ii) sliding-window teacher Distribution Matching via DMD~\cite{yin2024dmd, yin2024dmd2}, and (iii) SFT on real long clips.
Removing the DDT dual heads and naively training a single velocity predictor with both mean-seeking SFT and mode-seeking teacher alignment leads to the largest drop across all metrics, validating our motivation that the two objectives produce gradient interference when forced through one head.
Disabling sliding-window DMD degrades the model into SFT-only.
Removing SFT and relying only on sliding-window teacher alignment yields competitive motion but worse global consistency and overall quality, highlighting that a short-video teacher is inherently blind to minute-scale narrative structure and cannot substitute for long-video supervision.
Taken together, these results show that SFT provides the long-horizon coherence anchor, teacher distribution matching restores short-horizon fidelity, and the decoupled dual-head architecture is essential for combining them effectively.

\section{Conclusion}
\label{sec:conclusion}
\looseness=-1
We presented Mode Seeking meets Mean Seeking, a simple training paradigm for scaling video diffusion from seconds to minutes under scarce long-video data.
We use Decoupled Diffusion Transformer, which separates objectives with a shared long-context encoder and two heads: a mean-seeking SFT flow-matching head that learns minute-scale coherence from real long clips, and a mode-seeking sliding-window distribution-matching head that aligns student windows to a frozen short-video teacher via reverse-KL distribution matching, preserving short-horizon realism without requiring additional short-video supervision.
Across quantitative, qualitative, and ablation results, this decoupling closes the fidelity–horizon gap by jointly improving local sharpness/motion and long-range consistency.

\para{Limitations}
\looseness=-1
Our method improves the fidelity--horizon trade-off, but it does not solve all forms of long-horizon planning.
In failure cases, the model can satisfy local visual realism while misinterpreting compositional temporal instructions, such as changing a subject's action state~(e.g., walking to flying) or revisiting a previously seen landmark rather than generating a genuinely cyclic route.
This suggests that stronger event-level planning, explicit memory of entities/places, and evaluation protocols that test temporal causality and revisitation remain important future work.
We refer to Sec.~\ref{sec:limitations} for further discussion on limitations and future work.

\section*{Acknowledgements}
\looseness=-1
We thank Shuai Yang for the fruitful discussion and explanation regarding LongLive~\cite{yang2026longlive}.
We thank Zhen Li to provide support regarding the Sekai~\cite{li2025sekai} dataset.
We thank the NVIDIA Academic Grant Program for providing part of the GPU resources for this project.

\section*{Impact Statement}

This paper presents work whose goal is to advance the field of Machine
Learning. There are many potential societal consequences of our work, none
which we feel must be specifically highlighted here.

\bibliography{example_paper}
\bibliographystyle{icml2026}

\flushcolsend

\newpage
\appendix
\onecolumn
\section{Gemini Evaluation Prompt.}
\label{sec:gemini_eval_prompt}
\begin{quote}
{\small
\texttt{You are evaluating a video generation result.}

\texttt{Task: score the *semantic consistency* of the video on a 0-100 scale.}

\texttt{Semantic consistency means: objects, identities, attributes, and the overall scene remain coherent over time;}

\texttt{penalize sudden unrealistic changes, object identity swaps, content drift, or contradictions between frames.}

\texttt{IMPORTANT: account for whether the video is trivially consistent because it is static.}

\texttt{If the video is essentially a still image / frozen frame(s) with little-to-no motion or temporal change, do NOT give a high score.}

\texttt{In that case, assign a low score 
because it does not demonstrate temporal consistency under motion.}

\texttt{Return ONLY a single integer between 0 and 100 (inclusive). No other words.}
}
\end{quote}

\section{Additional Experiment on One-Minute Generation}
\label{sec:one_minute_generation}
\looseness=-1
To test whether the fidelity--horizon trade-off persists beyond the 30-second setting in the main paper, we additionally evaluate 60-second generation using the same automatic protocol as Sec.~\ref{sec:results}.
Because fully retraining the SFT-only baselines at this horizon requires a separate long-horizon tuning run, this additional evaluation focuses on rollout and context-extension baselines for which 60-second generation was available.
Thus, the table should be read as a one-minute stress test rather than a replacement for the complete 30-second benchmark in Tab.~\ref{tab:quantitative}.

\begin{table}[t]
\centering
\resizebox{0.9999\linewidth}{!}{
\begin{tabular}{@{} l | c | c | c | c | c | c | c }
\toprule
\multirow{2}{*}{Method} & Subject & Background & Motion & Dynamic & Aesthetic & Image & VLM \\
& Consistency $\uparrow$ & Consistency $\uparrow$ & Smoothness $\uparrow$ & Degree $\uparrow$ & Quality $\uparrow$ & Quality $\uparrow$ & Consistency $\uparrow$ \\
\midrule
CausVid~\cite{yin2025causvid} & \pf{0.9702} & \pf{0.9589} & 0.9771 & 0.5078 & \pt{0.5596} & 0.6264 & 41.82 \\
Self-Forcing~\cite{huang2025selfforcing} & 0.9516 & 0.9467 & 0.9784 & 0.4609 & 0.5524 & 0.6291 & 40.37 \\
InfinityRoPE~\cite{yesiltepe2026infinityrope} & \ps{0.9683} & 0.9568 & \pt{0.9810} & \pt{0.7891} & 0.5415 & \pt{0.6826} & \ps{66.08} \\
LongLive~\cite{yang2026longlive} & \pt{0.9679} & \ps{0.9572} & \ps{0.9827} & \ps{0.8203} & \ps{0.5704} & \ps{0.6908} & \pt{62.74} \\
\midrule
Ours & \underline{0.9676} & \pt{\underline{0.9569}} & \pf{\underline{0.9839}} & \pf{\underline{0.9297}} & \pf{\underline{0.5861}} & \pf{\underline{0.7015}} & \pf{\underline{73.08}} \\
\bottomrule
\end{tabular}
}
\caption{
\textbf{Additional 60-second quantitative comparison.}
The \colorbox{red!40}{first}, \colorbox{orange!50}{second}, and \colorbox{yellow!50}{third} values are highlighted within this 60-second stress test.
Our method achieves the strongest motion smoothness, dynamic degree, aesthetic quality, image quality, and VLM consistency, while remaining competitive on subject/background consistency.
}
\label{tab:one_minute_quantitative}
\end{table}

\looseness=-1
The one-minute results reinforce the interpretation of the quantitative results in Sec.~\ref{sec:results}.
CausVid and LongLive remain strong on frame-level subject/background consistency, but their lower dynamic degree and VLM consistency indicate more conservative temporal evolution.
In contrast, our model obtains the best dynamic degree~(0.9297), image quality~(0.7015), and VLM consistency~(73.08), suggesting that the decoupled mean-seeking and mode-seeking objectives preserve both local quality and long-horizon evolution at the longer generation horizon.

\section{User Study}
\label{sec:user_study}
\looseness=-1
We conducted a user study with 31 participants.
Each participant viewed randomized sets of baseline and ours videos: 25 sets at 30 seconds and 15 sets at 60 seconds.
Participants could scrub through each video and rated overall quality, global consistency, and static artifacts.
Scores are averaged over participants; higher is better for overall quality and global consistency, while lower is better for static artifacts.
As with Sec.~\ref{sec:one_minute_generation}, the 60-second setting excludes SFT-only baselines because they require separate retraining for this longer native horizon.

\begin{table}[t]
\centering
\resizebox{0.8\linewidth}{!}{
\begin{tabular}{@{} l l | c | c | c }
\toprule
Horizon & Method & Overall Quality $\uparrow$ & Global Consistency $\uparrow$ & Static Artifact $\downarrow$ \\
\midrule
\multirow{7}{*}{30 sec}
& LongSFT & 2.85 & 3.09 & \pt{2.11} \\
& MixedSFT & 2.66 & \ps{3.42} & \pf{1.64} \\
& CausVid~\cite{yin2025causvid} & 1.98 & 1.92 & 2.98 \\
& Self-Forcing~\cite{huang2025selfforcing} & 2.71 & 1.75 & 2.82 \\
& InfinityRoPE~\cite{yesiltepe2026infinityrope} & \ps{3.23} & 2.99 & 4.12 \\
& LongLive~\cite{yang2026longlive} & \pt{3.12} & \pt{3.27} & 4.03 \\
& Ours & \pf{\underline{3.68}} & \pf{\underline{3.59}} & \ps{\underline{1.86}} \\
\midrule
\multirow{5}{*}{60 sec}
& CausVid~\cite{yin2025causvid} & 2.06 & 1.90 & 4.08 \\
& Self-Forcing~\cite{huang2025selfforcing} & 2.69 & 1.98 & 3.91 \\
& InfinityRoPE~\cite{yesiltepe2026infinityrope} & \pt{3.11} & \pt{2.86} & \ps{3.03} \\
& LongLive~\cite{yang2026longlive} & \ps{3.24} & \ps{3.06} & \pt{3.22} \\
& Ours & \pf{\underline{3.59}} & \pf{\underline{3.41}} & \pf{\underline{2.08}} \\
\bottomrule
\end{tabular}
}
\caption{
\textbf{User study.}
The \colorbox{red!40}{first}, \colorbox{orange!50}{second}, and \colorbox{yellow!50}{third} values are highlighted, with the ordering reversed for static artifacts because lower is better.
Our method receives the highest overall quality and global consistency ratings for both 30-second and 60-second videos, and substantially reduces static artifacts relative to rollout-based baselines.
}
\label{tab:user_study}
\end{table}

\section{Further Discussion on Limitations and Future Work}
\label{sec:limitations}
\looseness=-1
We also note that our method is orthogonal to causal autoregressive methods~\cite{yin2025causvid, huang2025selfforcing}.
A natural follow-up is to use it as a base model for causal AR training, or, conversely, to distill our long-context bidirectional model into a causal sampler, which could be as simple as adding a causal attention mask during training.
Because our student is trained with substantially longer native temporal context than typical short-clip teachers, we expect it to extrapolate well to minute-scale~(and longer) generation, especially when combined with rollout-robustness and context-extension techniques such as Rolling Forcing~\cite{liu2026rollingforcing}, LongLive~\cite{yang2026longlive}, or longer-context positional embedding schemes~(e.g., InfinityRoPE~\cite{yesiltepe2026infinityrope}).
Finally, our 60-second supplemental benchmark is a stress test against available rollout/context-extension baselines, not a complete replacement for retrained SFT-only baselines at that horizon.
More broadly, the native long-context encoder provides a persistent history representation in the spirit of Genie-style models, and adding interaction/action conditioning on top of this representation is an especially promising direction.

\end{document}